\documentclass{article} % For LaTeX2e
\usepackage{iclr2026_conference,times}
\usepackage{amsmath}
\usepackage{booktabs} 
\usepackage{graphicx}
\usepackage[table]{xcolor}
\usepackage{multirow}
\usepackage{multicol}
\usepackage{makecell}
\usepackage{array}
\usepackage{graphicx}
\usepackage{caption}

\makeatletter
\newcommand\figcaption{\def\@captype{figure}\caption}
\newcommand\tabcaption{\def\@captype{table}\caption}
\makeatother

\usepackage{amsmath,amsfonts,bm}

\def\eqref#1{equation~\ref{#1}}
\def\1{\bm{1}}

\DeclareMathAlphabet{\mathsfit}{\encodingdefault}{\sfdefault}{m}{sl}
\SetMathAlphabet{\mathsfit}{bold}{\encodingdefault}{\sfdefault}{bx}{n}

\usepackage{hyperref}
\usepackage{url}

\title{FURINA: Free from Unmergeable Router via lINear Aggregation of mixed experts}

\author{Jiayi Han \\
Inspur Genersoft, Inspur Group\\
\texttt{hanjiayi@inspur.com} \\
\And
Liang Du, Yinda Chen \\
Interactive Entertainment Group\\ Tencent Inc. \\
\And
Xiao Kang\\
Shandong University\\
\And
Weiyang Ding \\
Fudan University \\
\texttt{dingwy@fudan.edu.cn} \\
\And
Donghong Han \\
Northeastern University, China
}

\iclrfinalcopy % Uncomment for camera-ready version, but NOT for submission.
\begin{document}

\maketitle

\begin{abstract}
The Mixture of Experts (MoE) paradigm has been successfully integrated into Low-Rank Adaptation (LoRA) for parameter-efficient fine-tuning (PEFT), delivering performance gains with minimal parameter overhead. However, existing MoE-LoRA methods suffer from a critical limitation: their reliance on discrete routers prevents integration of MoE components into the backbone model, resulting in persistent computational overhead and increased system complexity during inference.
To address this challenge, we propose \textbf{FURINA}, a novel \textbf{F}ree from \textbf{U}nmergeable \textbf{R}outer framework based on L\textbf{IN}ear \textbf{A}ggregation of experts. To the best of our knowledge, FURINA represents the first fully mergeable MoE-LoRA method that can be seamlessly reparameterized into the backbone model after training. This enables FURINA to function as a plug-and-play component within any LLM deployment framework--a capability where standard MoE-LoRA approaches fail--while maintaining equivalent learning capacity.
FURINA introduces a Mergeable Self-Routing mechanism that leverages angular similarity between inputs and adapter directional components to activate experts, which are then scaled by the shared magnitude vector. This design enables the output norm to naturally reflect expert importance, facilitating both router-free operation and seamless merging. The expert selection loss further enhances this behavior by encouraging sparsity and alignment with standard MoE activation patterns. Additionally, we incorporate a shared expert within the MoE-LoRA block to provide stable, foundational knowledge.
Extensive experiments across 9 benchmarks and 3 different LLMs demonstrate that FURINA significantly outperforms standard LoRA while matching or surpassing existing MoE-LoRA methods, all while eliminating their additional inference-time overhead. We plan to open-source our implementation upon publication.
\end{abstract}


\begin{figure}[htbp]
    \centering
    \includegraphics[width=0.7\linewidth]{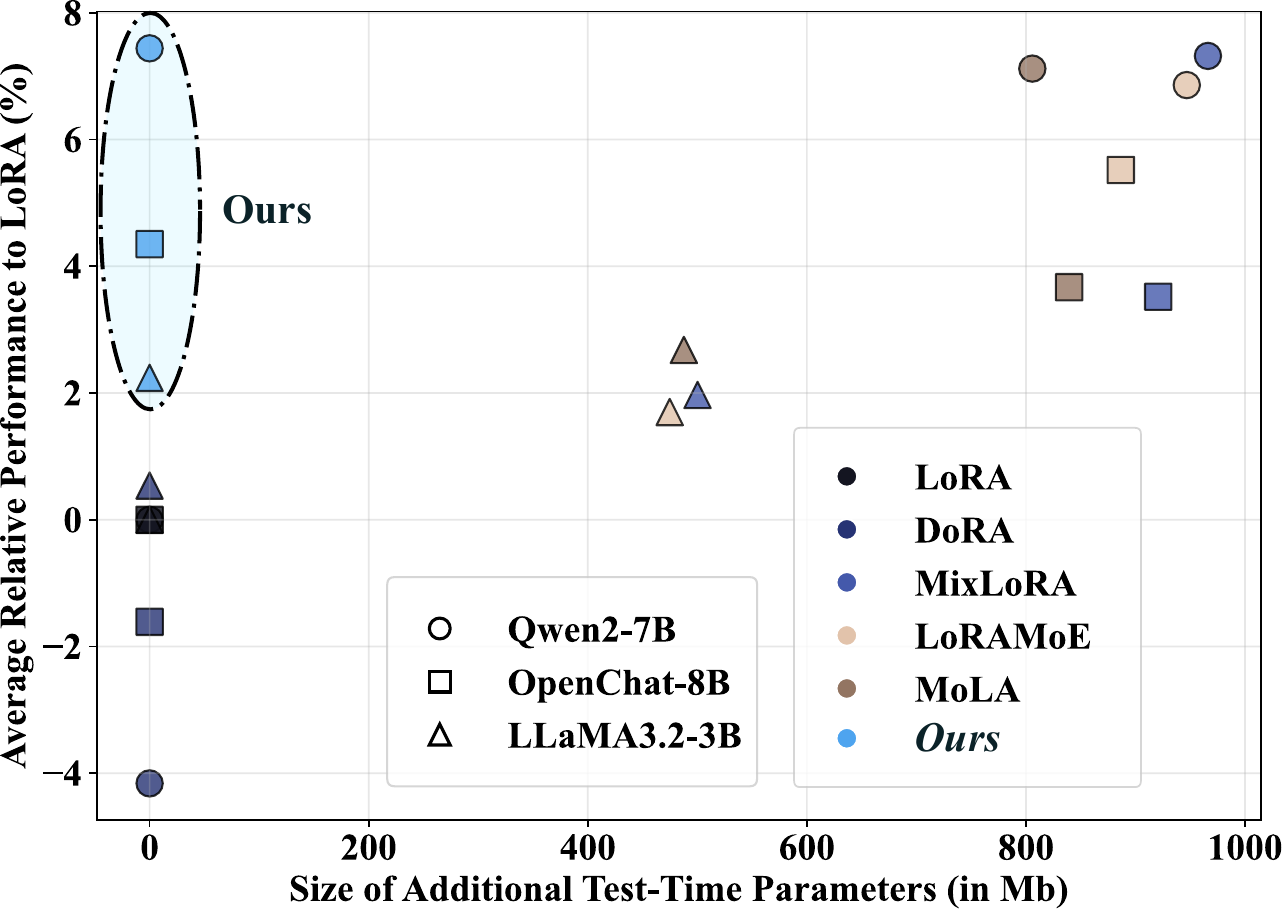}
    \caption{Relationship between additional test-time parameters and relative performance gain compared to standard LoRA. FURINA achieves comparable or superior performance gains to standard MoE-LoRA approaches while sharing the same architecture with standard LoRA during inference, resulting in up to 1.5$\times$ speedup.}
    \label{fig:fig1}
    \vspace{-5mm}
\end{figure}

\section{Introduction}
While Large Language Models (LLMs) have achieved remarkable success, effectively balancing their formidable performance with stringent resource constraints remains a significant challenge. Parameter-efficient fine-tuning (PEFT) methods, particularly Low-Rank Adaptation (LoRA), have emerged as a dominant solution, enabling effective adaptation of LLMs.
Inspired by the success of Mixture-of-Experts (MoE) architectures like Mixtral 8×7B (\cite{jiang2024mixtral}), recent work has sought to integrate MoE principles into LoRA-based PEFT to enhance model capacity within a fixed parameter budget. For instance, 
MixLoRA (\cite{li2024mixlora}) directly introduces the Mixtral-styled MoE into the LoRA approach. LoRAMoE (\cite{dou-etal-2024-loramoe}) proposes to alleviate knowledge on specific LoRA adapters, and SLIM (\cite{han-etal-2025-slim}) introduces identity layers and a dynamic merging strategy to leverage the pre-trained knowledge of the backbone model effectively.
Although integrating MoE with LoRA improves expressivity, it inherits a critical limitation: the introduction of an individual routing function prevents the adapters from being merged back into the base model. This fundamentally undermines a core advantage of standard LoRA, as the unmerged experts necessitate specialized, often inefficient, inference logic. Consequently, these methods suffer from limited support in high-performance inference frameworks like vLLM (\cite{vllm}) and incur unavoidable computational overhead, posing a major barrier to their practical deployment.
\begin{table}[htbp]
    \centering
    \caption{Comparison of the proposed FURINA with different fine-tuning strategies}
    \resizebox{0.7\linewidth}{!}{
    \begin{tabular}{cccc}
    \toprule
    Method & PEFT  & \makecell[c]{MoE During\\Training} & \makecell[c]{Zero Extra\\Test Time Cost} \\
    \midrule
    Full-SFT & $\times$ & $\times$ & $\checkmark$ \\
    \midrule
    LoRA (\cite{hulora})  & \multirow{2}[2]{*}{$\checkmark$} & \multirow{2}[2]{*}{$\times$} & \multirow{2}[2]{*}{$\checkmark$} \\
    DoRA (\cite{liudora})  &       &       &  \\
    \midrule
    MixLoRA (\cite{li2024mixlora}) & \multirow{4}[2]{*}{$\checkmark$} & \multirow{4}[2]{*}{$\checkmark$} & \multirow{4}[2]{*}{$\times$} \\
    LoRAMoE (\cite{dou-etal-2024-loramoe}) &       &       &  \\
    MoLA (\cite{gao-etal-2025-mola})  &       &       &  \\
    SLIM (\cite{han-etal-2025-slim}) &       &       &  \\
    \midrule
    FURINA (Ours)  & $\checkmark$ & $\checkmark$ & $\checkmark$ \\
    \bottomrule
    \end{tabular}%
    }
    \label{tab:comparison}
\end{table}

To overcome these limitations while preserving the performance benefits of MoE, we propose FURINA ({\bf F}ree from {\bf U}nmergeable {\bf R}outer via L{\bf IN}ear {\bf A}ggregation), a novel fully mergeable MoE-enhanced LoRA architecture which matches the performance of the standard MoE--LoRA methods while overcoming the deployment complexity and overhead.
The core of our approach is a Mergeable Self-Routing mechanism, which replaces the traditional router with three key components:
(1) Decoupled Learning of Direction and Magnitude: decouples the LoRA adapters by applying column-wise normalization to the weight matrices, isolating their directional components.
(2) Shared Learnable Magnitude Vector: introduces a single, shared magnitude vector that scales the outputs of all experts uniformly, ensuring the norm of an expert's output directly reflects its activation strength.
(3) Expert Selection Loss: employs a loss function that encourages sparse, divergent expert activation by maximizing the contribution of the most relevant experts.
Specifically, the input is first projected by the normalized LoRA matrices to produce normalized logits, then scaled by the shared magnitude vector. This design allows the norm of each expert's output to naturally represent its relevance to the input, enabling dynamic, router-free routing.
We also introduce a Shared Expert (SE) within the MoE--LoRA block. This expert provides foundational knowledge across the data corpus without introducing non-linearities that would prevent merging.
By integrating these modules, FURINA seamlessly transitions between two phases: during training, it operates as a full, capacity-enhanced MoE architecture; during inference, the experts and the shared expert are linearly aggregated and can be merged into a single LoRA adapter or directly into the backbone model, introducing zero overhead. A comparative summary of FURINA against full fine-tuning and other PEFT methods is provided in Tab.~\ref{tab:comparison}.

Our contributions are threefold:
\begin{enumerate}
    \item We propose FURINA, the first, to our knowledge, fully mergeable MoE--LoRA architecture. Unlike existing MoE-LoRA methods, FURINA can be seamlessly re-parameterized into a single LoRA adapter or directly into the backbone LLM after training. This ensures full compatibility with high-performance inference frameworks like vLLM and introduces \textbf{zero additional latency or complexity} during deployment.
    \item We introduce a novel Self-Routing mechanism that eliminates the need for a discrete router via three key innovations: (a) decoupling the learning of direction and magnitude in adapters, (b) a shared magnitude vector for uniform activation scaling, and (c) an expert selection loss that promotes specialization. We also propose a shared expert to mitigate the diminished output norm, preserving model capacity.
    \item We conduct extensive experiments on multiple LLMs and benchmarks. The results demonstrate that FURINA significantly enhances the performance of standard LoRA and achieves competitive or superior results compared to state-of-the-art non-mergeable MoE-LoRA methods, while eliminating the corresponding inference-time costs.
\end{enumerate}

\begin{figure}[tbp]
    \centering
    \includegraphics[width=1.\linewidth]{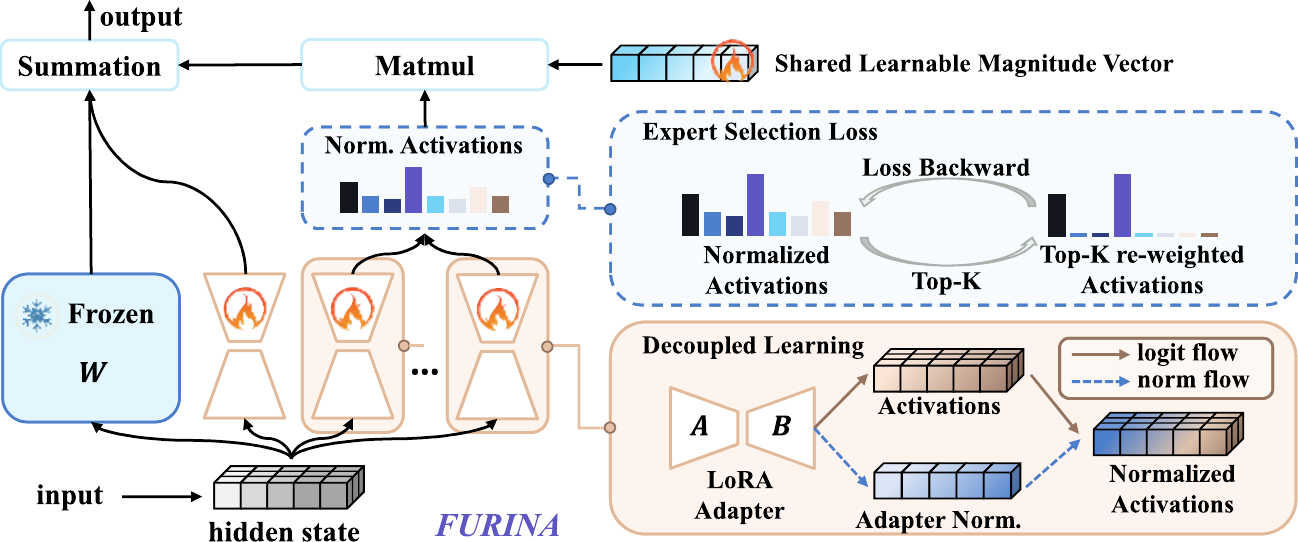}
    \caption{The overall framework of FURINA during training. LoRA weights are normalized, multiplied by the input hidden state, and scaled by a shared magnitude vector before aggregation with the backbone and shared expert outputs, while an expert selection loss sharpens activation sparsity.
    }
    \label{fig:framework}
\end{figure}

\section{Preliminaries}

\paragraph{Low-Rank Adaption} LoRA introduces trainable low-rank matrices to adapt large-scale pre-trained models efficiently. Typically applied to the multi-layer perceptron (MLP) layers, given a frozen weight matrix $W \in \mathbb{R}^{c_2 \times c_1}$ and an input hidden state $x \in \mathbb{R}^{c_1}$, the output of a LoRA-adapted layer is formulated as:
\begin{equation}
y = Wx + BAx,
\end{equation}
where $A \in \mathbb{R}^{r \times c_1}$ and $B \in \mathbb{R}^{c_2 \times r}$ are learnable low-rank matrices with rank $r \ll \min(c_1, c_2)$.

\paragraph{MoE--LoRA}
The MoE paradigm has been integrated with LoRA to enhance model capacity while maintaining parameter efficiency. This approach initializes $N$ distinct LoRA adapters ${(B_i, A_i)}_{i=1}^N$ and employs a router function $\phi$ to select a sparse combination of $K$ experts for each input.
The router is typically implemented as a trainable gating network. For input $x$, the routing weights are computed as
\begin{equation}
\phi(x)_i = \left\{
   \begin{array}{l}
       \frac{1}{Z}r(x)_i,i\in\mathop{\text{arg top}K}\limits_i\big(r(x)_i\big),\\
       0, \text{otherwise},
   \end{array}
\right.r(x) = \text{Softmax}(W_Gx),
\end{equation}
where $W_G \in \mathbb{R}^{N \times c_1}$ is a learnable projection matrix and $Z$ is a normalization constant. The final output becomes
\begin{equation}
   \begin{array}{l}
       y = Wx + \sum\limits_{i\in U}^N\phi(x)_iB_iA_ix, \quad U=\mathop{\text{arg top}K}\limits_j(r(x)_j),
   \end{array}
   \label{eq:moe}
\end{equation}
while MoE-LoRA enhances model capacity, it introduces a critical limitation: the discrete routing function $\phi(x)$ prevents the merging of adapters into the backbone model. The adapted weight $W' = W + \sum_i \phi(x)_i B_i A_i$ varies with each input $x$, necessitating separate execution of router and experts during inference. This significantly increases implementation complexity and computational overhead compared to standard LoRA.
The proposed FURINA framework addresses this fundamental limitation while preserving the capacity benefits of MoE-LoRA, enabling both enhanced expressivity and efficient deployment.

\section{Method}

\subsection{Overview of the proposed FURINA}
As detailed in the previous section, the MoE-enhanced LoRA architecture employs a router to dynamically activate specific LoRA adapters. To emulate this behavior, the proposed FURINA incorporates three key components: (1) Decoupled learning of direction and magnitude for LoRA adapters, (2) Shared learnable magnitude vector for uniform activation scaling, and (3) Expert selection loss that encourages divergent activation like MoE. As depicted in Fig.~\ref{fig:framework}, for each LoRA adapter, we simultaneously calculate the activation of the input hidden state and the norm of the adapter. The activations are multiplied by the reciprocal of the adapter norm for normalization. To avoid the uniform activation, the expert selection loss maximizes the activation share of the top-K LoRA experts. We also introduce shared experts to capture the shared knowledge across FURINA experts. Without a router, the shared experts could be combined with the backbone.

\subsection{Mergeable Self-Routing mechanism}
Mergeable Self-routing is the most important mechanism of FURINA, enabling the merging of LoRA experts. 
Since $W$ is frozen, we focus mainly on the adapted part of the output hidden state. For simplicity, we denote the adapted part of the MLP layer for MoE--LoRA as follows:
\begin{equation}
    \begin{array}{l}
        \Delta y = y-Wx=\sum_i\phi(x)_iB_iA_ix.
    \end{array}
    \label{eq:deltay}
\end{equation}
The simplest way to eliminate the routers is to assign the same weight to all experts:
\begin{equation}
    \begin{array}{l}
        \Delta y = \sum\limits_{i}^NB_iA_ix.
    \end{array}
    \label{eq:naive-no-router}
\end{equation}
However, this formulation will degrade the MoE of LoRA to the naive LoRA, where $B'=(B_1, B_2, \ldots,B_N)$ and $A'=(A_1^T,A_2^T,\ldots,A_N^T)^T$. 

\noindent\textbf{Decoupled Learning }
To solve the aforementioned issue, we need to review the original MoE of LoRA. 
Denote the full-size weight matrix of the $i_{th}$ LoRA adapter as $W_i=B_iA_i$, and its column-wise norm could be calculated as follows:
\begin{equation}
    \begin{array}{l}
        d_i = W_ie, \quad e=\underbrace{[1,1,\ldots,1]^T}_{\times c_1}.
    \end{array}
\end{equation}
Denote $\hat{W}_i=\text{diag}(d_i+\epsilon)^{-1}W_i$, in which $\epsilon$ is a small positive number to prevent division of $0$, and $\text{diag}$ expands a vector to the corresponding diagonal matrix. The formulation of MoE--LoRA in Eq.~\ref{eq:deltay} could be reformulated as follows:
\begin{equation}
    \begin{array}{l}
        \Delta y=\sum\limits_i^N\underbrace{\text{diag}(d_i+\epsilon)}_{\text{magnitude}}\cdot\underbrace{\phi(x)_i\hat{W}_i\hat{x}}_{\text{reweighted similarity}}\cdot\Vert x\Vert,
    \end{array}
    \label{eq:deltay-v2}
\end{equation}
in which $\hat{x}=x/\Vert x\Vert$. Similarly, after eliminate the router, Eq.~\ref{eq:naive-no-router} could be reformulated as follows:
\begin{equation}
    \begin{array}{l}
        \Delta y=\sum\limits_i^N\underbrace{\text{diag}(d_i+\epsilon)}_{\text{magnitude}}\cdot\underbrace{\hat{W}_i\hat{x}}_{\text{similarity}}\cdot\Vert x\Vert.
    \end{array}
    \label{eq:lora_reformed}
\end{equation}
Different from Eq.~\ref{eq:deltay-v2}, eliminating the router creates a problematic coupling between the magnitude and similarity terms. The similarity calculation embeds the inverse of the magnitude without any independent operations.
Consequently, when the similarity term is multiplied by the magnitude term, they cancel each other out due to the inverse relationship.  This cancellation nullifies the intended effect entirely and disables measuring input-to-adapter similarity via the output norm.
To address this, we decouple the magnitude and the similarity by replacing the magnitude with a learnable vector $v_i\in\mathcal{R}^{c_2}$, and approximate the MoE of LoRA adapters as follows:
\begin{equation}
    \begin{array}{l}
        \Delta y=\sum\limits_{i}^N \text{diag}(v_i)(\hat{W}_i\hat{x}\Vert x\Vert)=\sum\limits_{i}^N\text{diag}(v_i)(\hat{W}_ix).
    \end{array}
\end{equation}

\noindent\textbf{Shared Magnitude Vector } 
Although decoupled learning addresses the issue of magnitude-similarity cancellation, the following issue still exists: the output norm of an expert could be large even if the similarity term is small (indicating the expert should not be activated), when the magnitude $v_i$ is large. To address this, we introduce a shared magnitude vector $v$ for all experts.
It is worth noting that we do not need to calculate the full-size $\hat{W}$ and $\text{diag}(v)$ during training.
The output could be further reformulated as follows:
\begin{equation}
    \begin{array}{l}
        \Delta y=\sum\limits_{i}^N\text{diag}(v)(\hat{W}_ix)=\sum\limits_{i=1}^N \Big(v\otimes \frac{1}{d_i+\epsilon}\Big)\otimes B_iA_ix,
    \end{array}
\end{equation}
in which $\otimes$ represents element-wise multiplication. Note that $\tilde{B}_i$ is only calculated once per batch.

\paragraph{Shared Experts}
Unlike standard MoE--LoRA methods that normalize routing weights to enforce exactly $K$ active experts, cross-expert re-weighting is not applicable in our approach. In extreme cases, the output of the MoE may approach zero, thereby limiting its learning capacity. To mitigate this issue, we introduce a shared expert component to FURINA. Specifically, the incremental output $\Delta y$ with shared expert is calculated as follows:
\begin{equation}
    \begin{array}{l}
        \Delta y=\sum\limits_{i=1}^{N_{SE}}B_iA_ix+\sum\limits_{j=N_{SE}+1}^{N}\text{diag}(v)\hat{W}_ix.
    \end{array}
    \label{eq:FURINAse}
\end{equation}

\subsection{Training Objectives}

The training objective consists of two parts: (1) the supervised fine-tuning (SFT) loss, and (2) the expert selection loss. The SFT loss, similar to the prior approaches, is defined as follows:
\begin{equation}
    \begin{array}{l}
        \mathcal{L}_{\text{SFT}} = \frac{1}{|y|}\sum\limits_t\text{CE}(f(y_t|x,y_{<t})),
    \end{array}
\end{equation}
in which $\text{CE}$ represents the cross-entropy loss.

\noindent\textbf{Expert Selection Loss }We propose the expert selection loss to encourage the self-routing of LoRA adapters to approximate the function of routers.
Specifically, denote the activations of the $i_{th}$ LoRA adapter (apart from the shared experts, if any) as $\textbf{a}_i$.
To encourage divergent activation per token, we introduce the divergence loss as follows:
\begin{equation}
    \begin{array}{l}
        \mathcal{L}_{\text{div}} = -\text{log}\left(\frac{\sum\limits_{i\in \mathcal{S}}|\text{sum}(\textbf{a}_i)|}{\sum\limits_j|\text{sum}(\textbf{a}_j)|}\right), \mathcal{S} = \mathop{\text{arg top}K}\limits_k\left(\Big|\sum\limits\limits_j\textbf{a}_{k,j}\Big|\right),\textbf{a}_i = \hat{W}_ix
    \end{array},
\end{equation}
in which $\mathcal{S}$ demonstrates the indices of the selected experts.
Moreover, we also introduce the balance loss of expert selection. Specifically, given a batch of logits $\mathcal{X}\in\mathbb{R}^{B,T,N}$ in which $B, T, N$ demonstrate the batch size, number of tokens per sample, and the total number of experts. First, we calculate the activation frequency of each expert:
\begin{equation}
    \begin{array}{l}
        \mathcal{F}_i=\Big(\sum_u\sum_v \mathbb{I}(i\in \mathcal{S}_{u,v})\Big)\Big/T, \mathcal{P}_i=\sum_u\sum_v\left\vert\mathcal{X}_{u,v,i}\right\vert\Big/T
    \end{array},
\end{equation}
in which $\mathcal{S}_{u,v}$ represents the selected expert of token $\mathcal{X}_{u,v,:}$, $\mathbb{I}(\cdot)=1$ if the input condition is ``True'', otherwise it equals to $0$. Then the balance loss could be calculated as follows:
\begin{equation}
    \begin{array}{l}
        \mathcal{L}_{\text{bal}} = N\times\sum_i\mathcal{F}_i\mathcal{P}_i.
    \end{array}
\end{equation}

Denote it as $\mathcal{L}_{\text{bal}}$, given a certain batch of input, the expert selection loss $\mathcal{L}_{sel}$ is defined as their summation. The overall training objective is defined as follows:
\begin{equation}
    \begin{array}{l}
        \mathcal{L} = \mathcal{L}_{\text{SFT}} + \alpha\mathcal{L}_{\text{sel}}, \quad\mathcal{L}_{\text{sel}}=\mathcal{L}_{\text{div}} + \mathcal{L}_{\text{bal}},
    \end{array}
\end{equation}
in which $\alpha$ represents the loss coefficient, which is set to $0.01$ in our work.

\subsection{Merging of Experts}
During inference, unlike the standard MoE--LoRA approaches, the multiple LoRA adapters of the proposed FURINA could be merged without loss of information. 
Without loss of generality, we start from the full FURINA to merge all the experts into one LoRA adapter, and furthermore, into the backbone network. 
First, we re-write Eq.~\ref{eq:FURINAse} as follows:
\begin{equation}
    \begin{aligned}
        \Delta y&=\sum\limits_{i=1}^{N_{SE}}B_iA_ix+\sum\limits_{j=N_{SE}+1}^{N}\text{diag}(v)\hat{W}_jx\\
        &=\sum\limits_{i=1}^{N_{SE}}B_iA_ix+\sum\limits_{j=N_{SE}+1}^{N}\text{diag}(v)\otimes \frac{1}{d_j+\epsilon}(B_jA_jx)\\
        &=\sum\limits_{i=1}^{N_{SE}}B_iA_ix+\sum\limits_{j=N_{SE}+1}^{N}\text{diag}\Bigg(v\otimes \frac{1}{d_j+\epsilon}\Bigg)  B_jA_jx\\
        &=\sum\limits_{i=1}^{N_{SE}}B_iA_ix+\sum\limits_{j=N_{SE}+1}^{N} \tilde{B_j}A_jx.\\
    \end{aligned}
\end{equation}
Then we could merge all these LoRA adapters and the backbone as follows: 
\begin{equation}
\left\{
    \begin{array}{l}
         \textbf{B}=(B_1,B_2,\ldots,B_{N_{SE}},\tilde{B}_{N_{SE}+1},\ldots,\tilde{B}_{N}),\\
         \textbf{A}=(A_1^T,A_2^T,\ldots,A_N^T)^T.
    \end{array}
\right.
\end{equation}
Then the output could be formulated as:
\begin{equation}
    \begin{array}{l}
        y+\Delta y=Wx+\textbf{B}\textbf{A}x=(W+\textbf{B}\textbf{A})x.
    \end{array}
\end{equation}
For FURINA without shared experts, $N_{SE}=0$, $\textbf{B}$ could be reformulated as follows:
\begin{equation}
    \begin{array}{l}
         \textbf{B}=(\tilde{B}_1,\tilde{B}_2,\ldots,\tilde{B}_{N}).
    \end{array}
\end{equation}
\section{Experiments}
\subsection{Implementation Details}
 For each model, we leverage 7 benchmarks to evaluate the capacity of the PEFT methods, including CSQA (\cite{talmor2019commonsenseqa}), HellaSwag (\cite{zellers2019hellaswag}), Winogrande (\cite{sakaguchi2021winogrande}), ARC-c and ARC-e (\cite{clark2018think}), OBQA (\cite{mihaylov2018can}), and BoolQ (\cite{clark-etal-2019-boolq}). Details of the benchmarks are provided in the Appendix. 
 Three different LLMs are included in our experiment: Qwen2-7B (\cite{yang2024qwen2technicalreport}), OpenChat-8B (\cite{wangopenchat}), and LLaMA3.2-3B, covering different architectures and model scales.
 Following the setting in MixLoRA (\cite{li2024mixlora}), the learning rate, the rank of each LoRA adapter $r$, the number of experts $N$, and the number of activated experts $K$ are set to $2\times 10^{-4}$, 16, 8, and 2, respectively. 
 For our proposed FURINA method, we include one shared expert alongside the self-routed experts (maintaining eight experts in total).
 All experiments are conducted on Nvidia GPUs.
 
\subsection{Comparison with SOTA Approaches}
We compare FURINA against state-of-the-art PEFT methods, including mergeable baselines (LoRA, DoRA) and non-mergeable MoE-LoRA approaches (MixLoRA, LoRAMoE, MoLA). We measure Time to First Token (TTFT) and latency using Llama3.2-3B. Mergeable methods (LoRA, DoRA, FURINA) are evaluated using vLLM+EvalScope, while non-mergeable methods use MoE-PEFT. For a fair comparison under equivalent parameter budgets, we set LoRA and DoRA rank to 128, matching the total rank of MoE methods. 
As shown in Tab.~\ref{tab:main}, FURINA significantly improves upon standard LoRA (+4.8\% average gain) and achieves competitive performance with MoE-LoRA methods. Meanwhile, FURINA maintains inference latency equivalent to standard LoRA, with detailed settings and results in Appendix~\ref{appendix:time}.
\begin{table}[htbp]
  \centering
  \caption{Comparison with LoRA-styled PEFT approaches on downstream tasks. We employ \textcolor[rgb]{ .306, .643, .918}{\bf blue} and {\bf bold} to indicate the best and second-best results for each model and the average performance. $\dag$ represents FURINA without shared experts. $^{\ddag}$ represents that the method is not applicable to vLLM, thus evaluated on the MoE-PEFT framework. FURINA achieves competitive performance with standard MoE--LoRA methods with much less computational overhead.}
  \resizebox{1.0\linewidth}{!}{
    \begin{tabular}{c|cccc|cc}
    \toprule
    Method & AVG & OpenChat-8B & Llama3.2-3B & Qwen2-7B & \makecell[c]{TTFT\\(ms, $\downarrow$)} & \makecell[c]{Latency\\(ms, $\downarrow$)}\\
    \midrule
    \multicolumn{7}{c}{Single Adapter LoRA} \\
    \midrule
    LoRA  & 78.8 & 80.6  & 77.4  & 78.5  & \multirow{2}{*}{\textcolor[rgb]{ .306,  .643,  .918}{\bf $\approx$ 10}} & \multirow{2}{*}{\textcolor[rgb]{ .306,  .643,  .918}{\bf $\approx$ 800}} \\
    DoRA  & 77.1 & 79.0  & 77.9  & 74.3  &  &  \\
    \midrule
    \multicolumn{7}{c}{Standard MoE of LoRA} \\
    \midrule
    MixLoRA & 83.1 & 84.1  & 79.3  & {\bf 85.8} & $\approx$550$^{\ddag}$  & $\approx$9000$^{\ddag}$ \\
    LoRAMoE & {\bf 83.5} & \textcolor[rgb]{ .306,  .643,  .918}{\bf 86.1} & 79.1  & 85.3  & $\approx$450$^{\ddag}$ & $\approx$6000$^{\ddag}$\\
    MoLA  & 83.4 & 84.3  & \textcolor[rgb]{ .306,  .643,  .918}{\bf 80.3} & 85.6  & $\approx$700$^{\ddag}$ & $\approx$14500$^{\ddag}$ \\
    \midrule
    \multicolumn{7}{c}{Fully Mergeable MoE of LoRA} \\
    \midrule
    {\bf FURINA$^{\dag}$ (Ours)} & 81.1 & \textcolor[rgb]{ .306,  .643,  .918}{\bf 86.1} & 73.9  & 83.3      & \multirow{2}{*}{\textcolor[rgb]{ .306,  .643,  .918}{\bf $\approx$ 10}} & \multirow{2}{*}{\textcolor[rgb]{ .306,  .643,  .918}{\bf $\approx$ 800}} \\
    {\bf FURINA (Ours)} & \textcolor[rgb]{ .306,  .643,  .918}{\bf 83.6} & {\bf85.3}  & {\bf 79.7} & \textcolor[rgb]{ .306,  .643,  .918}{\bf 85.9} &  &  \\
    \bottomrule
    \end{tabular}%
    }
  \label{tab:main}%
\end{table}%

\subsection{Compatibility with LoRA Variations}

We also evaluate the compatibility of FURINA with existing LoRA variations, including LoRA+ (\cite{hayou2024lora+}) and rsLoRA (\cite{kalajdzievski2023rank}). For LoRA+, we maintain a learning rate ratio of 5.0 between matrices $B$ and $A$. As shown in Tab.~\ref{tab:lora-modify}, FURINA is consistently compatible with these variants, achieving superior performance compared to standard MoE-LoRA approaches.

\begin{table}[htbp]
    \centering
    \caption{Compatibility of FURINA with LoRA variations}
    \begin{tabular}{ccccc}
    \toprule
        Method   &   OpenChat-8B   &   Llama3.2-3B   &   Qwen2-7B   &  AVG   \\
    \midrule
        LoRA     & 80.6     & 77.4     & 78.5     & 78.8  \\
        \midrule
        LoRAMoE  & 86.1     & 79.1     & 85.3     & 83.5  \\
        SLIM    & 87.4      & 79.2        & 85.9        & 84.2     \\
        \midrule
        {\bf FURINA} & 85.3     & 79.7     & 85.9     & 83.6  \\
        {\bf FURINA w/ LoRA+} & 87.2     & \textcolor[rgb]{ .306,  .643,  .918}{\bf 81.3}     & \textcolor[rgb]{ .306,  .643,  .918}{\bf 86.9}     & 85.1  \\
       {\bf  FURINA w/ rsLoRA}  & \textcolor[rgb]{ .306,  .643,  .918}{\bf 87.6}     & 81.2     &  \textcolor[rgb]{ .306,  .643,  .918}{\bf 86.9}    & \textcolor[rgb]{ .306,  .643,  .918}{\bf 85.2}\\
    \bottomrule
    \end{tabular}
    \label{tab:lora-modify}
\end{table}

\subsection{Ablation study}
\paragraph{Ablations on the Main Modules}

We conduct the ablation study on the main modules proposed in our work. We include Qwen2-7B and LLaMA3.2-3B with all benchmarks, apart from the HellaSwag (because of its size), in the main ablation study. ``Decoupled Learning'' demonstrates the column normalization of the LoRA adapters, and ``Shared Magnitude Vector'' represents the introduction of the shared magnitude vector $v$ after normalization. The result shown in Tab.~\ref{tab:main_abl} demonstrates that normalizing the LoRA adapters significantly improves the model performance by 3.4\%, representing the importance of balanced effectiveness of each expert. The introduction of shared experts and shared magnitude vector also boosts the model performance by 1.4\%, demonstrating the benefit of learning mutual foundation knowledge. Further mimicking the activation pattern of standard MoE by $\mathcal{L}_{\text{sel}}$ boosts the model by 0.3\%, indicating the importance of emulating the top-K pattern of MoE.
\begin{table}[htbp]
  \centering
  \caption{Ablation study of the main modules. The performance (Perf.) is the average of Qwen2-7B and LLaMA-3.2-3B.
}
    \begin{tabular}{ccccc}
    \toprule
    
    \makecell[c]{Decoupled \\Learning} & \makecell[c]{Shared \\Experts} & \makecell[c]{Shared \\Magnitude Vector} & $\mathcal{L}_{\text{sel}}$& Perf (\%) \\
    \midrule
    $\times$         & $\times$         & $\times$         & $\times$     &  75.9\\
    $\checkmark$     & $\times$         & $\times$         & $\times$     & 79.3  \\
    $\checkmark$     & $\checkmark$     & $\times$         & $\times$     &  80.2\\
    $\checkmark$     & $\checkmark$     & $\checkmark$     & $\times$     &  80.7\\
    $\checkmark$     & $\checkmark$     & $\checkmark$     & $\checkmark$ & {\bf 81.0} \\
    \bottomrule
    \end{tabular}%
  \label{tab:main_abl}%
\end{table}%

\begin{figure}
    \centering
    \includegraphics[width=.95\linewidth]{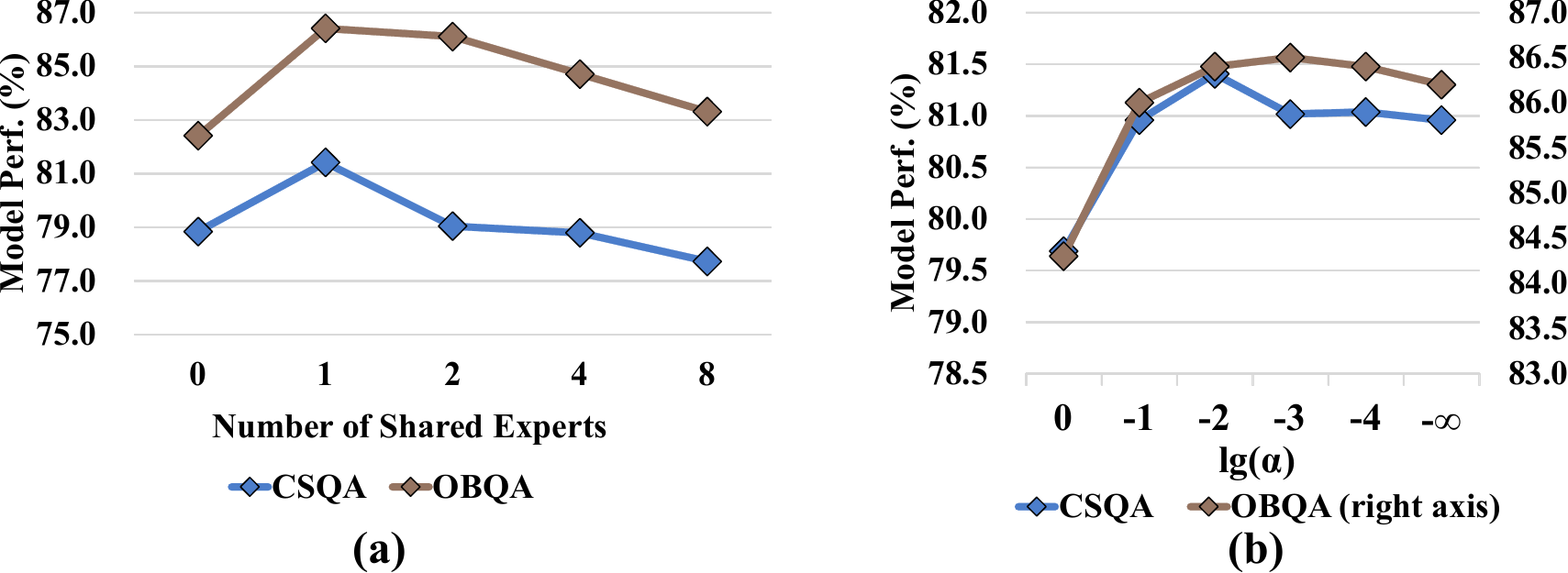}
    \caption{The effect of the number of shared experts (a) and the scale of the loss coefficient $\alpha$ (b)}
    \label{fig:abl}
\end{figure}

\paragraph{Effect of Number of Shared Experts}
We conduct an experiment on the OBQA and CSQA datasets to validate the influence of the number of shared experts. We average the performance of LLaMA-3.2-3B and Qwen2-7B. The result is demonstrated in Fig.~\ref{fig:abl}(a). Validation across different LLMs demonstrates that setting the number of shared experts to 1 is sufficient. Further increasing it may result in model performance degradation and convergence to the original LoRA approach.

\paragraph{Effect of Loss Coefficient}
We also conduct an experiment to validate the influence of the loss coefficient $\alpha$. The result is demonstrated in Fig.~\ref{fig:abl}(b). A large $\alpha$ results in over-focus on the imitation of the expert selection patterns of MoE models and limits the learning capacity of FURINA. 
On the contrary, a minor $\alpha$ cannot guide the LoRA adapters of the MoE to mimic the selective expert activation pattern of standard MoE architectures. Experiments demonstrate that setting $\alpha$ to the range of $[1e-2, 1e-3]$ could achieve an optimal trade-off for these two factors.

\paragraph{Comparison of Shared Magnitude Vector with Decoupling LoRA Adapters} To validate the utilization of the shared magnitude vector instead of simply decouple the direction and magnitude of the LoRA adapters, we conduct an experiment on the OBQA and CSQA datasets and the LLaMA3.2-3B model. The results are shown in Tab.~\ref{tab:decouple}. Utilizing the shared magnitude vector achieves higher performance with fewer trainable parameters. This may be because, although utilizing different magnitude vectors for the adapters increases the number of trainable parameters, it results in an imbalance in the activation of the adapters.
\begin{table}[htbp]
    \centering
    \caption{Comparison of shared magnitude vector with decoupling of LoRA direction and magnitude}
    \begin{tabular}{cccc}
    \toprule
        Method                  &   OBQA   &  CSQA  & AVG  \\
    \midrule
        Decoupling              &   82.4   &  78.5 & 80.4 \\
        {\bf Shared Mag. Vec. (Ours)} &   {\bf 84.0}   &  {\bf78.5} & {\bf81.2} \\
    \bottomrule
        
    \end{tabular}
    
    \label{tab:decouple}
\end{table}

\paragraph{Evaluation on GSM8K and HumanEval Datasets}
To validate the model capacity on reasoning tasks, we also conduct an experiment on the GSM8K (\cite{cobbe2021trainingverifierssolvemath}) and HumanEval (\cite{chen2021evaluating}) datasets. We keep the hyperparameters as in the main experiments, and all models are fine-tuned for 1k steps. For the HumanEval dataset, the model is trained on the CodeAlpaca (\cite{codealpaca}) dataset, and the metric is set to Pass@1. The Evalscope framework is adopted for evaluation. The results in Tab.~\ref{tab:gsm8k} demonstrate that, compared to LoRA, the proposed FURINA could significantly boost the model performance, indicating its generalizability.
\begin{table}[htbp]
\caption{Comparison of LoRA and FURINA on GSM8K and HumanEval datasets }
    \centering
    \begin{tabular}{ccccc}
    \toprule
    Method           & OpenChat-8B &   Llama-3.2-3B  &   Qwen2-7B    \\
    \midrule
    LoRA                & 61.1        &   47.8          & 62.5         \\
    {\bf FURINA (Ours)} & {\bf65.1}   &   {\bf53.4}     & {\bf 65.7}          \\
    \bottomrule
    \end{tabular}
    \label{tab:gsm8k}
\end{table}
\section{Related works}

\paragraph{LoRA-style PEFT}
LoRA (\cite{hulora}) has emerged as a prominent method for PEFT of large language models. Unlike prompt tuning or adapter-based approaches, LoRA's key advantage lies in its mergeability: after training, the low-rank adapters can be consolidated into the original pre-trained weights, introducing zero additional inference latency. 
DoRA (\cite{liudora}) decouples weight updates into magnitude and direction components, while rsLoRA (\cite{kalajdzievski2023rank}) introduces a scaling mechanism to improve training stability. \cite{hayou2024lora+} identifies an imbalance in learning dynamics between the $A$ and $B$ matrices and proposes differentiated learning rates to address it.
Recently, several works have integrated mixture-of-experts (MoE) architectures with LoRA-style PEFT. MixLoRA (\cite{li2024mixlora}) incorporates the sparse MoE structure from Mixtral into LoRA, and MoLA (\cite{gao-etal-2025-mola}) advocates for layer-wise expert configurations with dynamic expert counts. SLIM (\cite{han-etal-2025-slim}) further enhances this approach by blending identity mappings with LoRA adapters to better preserve pre-trained knowledge. A fundamental limitation of these MoE-LoRA methods, however, is their inability to merge experts into the base model post-training, resulting in persistent inference overhead and increased deployment complexity compared to standard LoRA.

\paragraph{Mixture of Experts Architecture}
MoE has gained significant traction for scaling large language models efficiently. Mixtral 8×7B (\cite{jiang2024mixtral}) stands as the first widely adopted open-source MoE LLM, demonstrating that sparse expert activation can substantially improve parameter efficiency without compromising performance. DeepSeek MoE (\cite{dai2024deepseekmoe}) introduces shared experts that remain active across all inputs to capture common knowledge, thereby enhancing the model's representational capacity. DeepSeek-V3 (\cite{liu2024deepseek}) further increases the number of shared experts and incorporates a loss-free balancing mechanism to improve training stability.
Recent innovations in MoE architectures also explore MoE frameworks for improved computational efficiency. \cite{jin2025moe} proposes incorporating non-computational experts—such as identity or constant-output layers—into the MoE framework, reducing inference costs while maintaining model performance. \cite{lv2025autonomyofexperts} proposes to remove the routers from MoE, but still retains the non-linear cross-expert operations that prevent merging of experts. 
\section{Conclusion}
In this work, we propose FURINA, a novel fully mergeable MoE--LoRA framework that overcomes the deployment complexity and overhead of the standard MoE--LoRA while matching and even surpassing their learning capacity.
FURINA introduces a Mergeable Self-Routing mechanism with three key components: (1) Decoupled learning of adapter direction and magnitude, (2) Shared magnitude vector for uniform scaling, and (3) Expert selection loss that promotes sparse expert activation. These elements collectively ensure that each expert's output norm reflects its relevance to the input, 
enabling mergeable routing.
We also incorporate shared experts within the MoE--LoRA block that provides essential foundational knowledge so that the other experts can focus on specific tasks.
Comprehensive experiments on 9 datasets and 3 LLMs demonstrate that FURINA significantly enhances standard LoRA performance while achieving competitive results compared to SOTA MoE-LoRA methods, while overcoming their complexity and overhead during deployment, and plug-and-play with mainstream deployment frameworks.

\section*{Acknowledgment}
This work was supported by the Shandong Provincial Natural Science Foundation (No. ZR2024QF128).

\bibliography{iclr2026_conference}

\begin{thebibliography}{24}
\providecommand{\natexlab}[1]{#1}
\providecommand{\url}[1]{\texttt{#1}}
\expandafter\ifx\csname urlstyle\endcsname\relax
  \providecommand{\doi}[1]{doi: #1}\else
  \providecommand{\doi}{doi: \begingroup \urlstyle{rm}\Url}\fi

\bibitem[Chaudhary(2023)]{codealpaca}
Sahil Chaudhary.
\newblock Code alpaca: An instruction-following llama model for code generation.
\newblock \url{https://github.com/sahil280114/codealpaca}, 2023.

\bibitem[Chen et~al.(2021)Chen, Tworek, Jun, Yuan, de~Oliveira~Pinto, Kaplan, Edwards, Burda, Joseph, Brockman, Ray, Puri, Krueger, Petrov, Khlaaf, Sastry, Mishkin, Chan, Gray, Ryder, Pavlov, Power, Kaiser, Bavarian, Winter, Tillet, Such, Cummings, Plappert, Chantzis, Barnes, Herbert-Voss, Guss, Nichol, Paino, Tezak, Tang, Babuschkin, Balaji, Jain, Saunders, Hesse, Carr, Leike, Achiam, Misra, Morikawa, Radford, Knight, Brundage, Murati, Mayer, Welinder, McGrew, Amodei, McCandlish, Sutskever, and Zaremba]{chen2021evaluating}
Mark Chen, Jerry Tworek, Heewoo Jun, Qiming Yuan, Henrique~Ponde de~Oliveira~Pinto, Jared Kaplan, Harri Edwards, Yuri Burda, Nicholas Joseph, Greg Brockman, Alex Ray, Raul Puri, Gretchen Krueger, Michael Petrov, Heidy Khlaaf, Girish Sastry, Pamela Mishkin, Brooke Chan, Scott Gray, Nick Ryder, Mikhail Pavlov, Alethea Power, Lukasz Kaiser, Mohammad Bavarian, Clemens Winter, Philippe Tillet, Felipe~Petroski Such, Dave Cummings, Matthias Plappert, Fotios Chantzis, Elizabeth Barnes, Ariel Herbert-Voss, William~Hebgen Guss, Alex Nichol, Alex Paino, Nikolas Tezak, Jie Tang, Igor Babuschkin, Suchir Balaji, Shantanu Jain, William Saunders, Christopher Hesse, Andrew~N. Carr, Jan Leike, Josh Achiam, Vedant Misra, Evan Morikawa, Alec Radford, Matthew Knight, Miles Brundage, Mira Murati, Katie Mayer, Peter Welinder, Bob McGrew, Dario Amodei, Sam McCandlish, Ilya Sutskever, and Wojciech Zaremba.
\newblock Evaluating large language models trained on code, 2021.

\bibitem[Clark et~al.(2019)Clark, Lee, Chang, Kwiatkowski, Collins, and Toutanova]{clark-etal-2019-boolq}
Christopher Clark, Kenton Lee, Ming-Wei Chang, Tom Kwiatkowski, Michael Collins, and Kristina Toutanova.
\newblock {B}ool{Q}: Exploring the surprising difficulty of natural yes/no questions.
\newblock In Jill Burstein, Christy Doran, and Thamar Solorio (eds.), \emph{Proceedings of the 2019 Conference of the North {A}merican Chapter of the Association for Computational Linguistics: Human Language Technologies, Volume 1 (Long and Short Papers)}, pp.\  2924--2936, Minneapolis, Minnesota, June 2019. Association for Computational Linguistics.
\newblock \doi{10.18653/v1/N19-1300}.
\newblock URL \url{https://aclanthology.org/N19-1300}.

\bibitem[Clark et~al.(2018)Clark, Cowhey, Etzioni, Khot, Sabharwal, Schoenick, and Tafjord]{clark2018think}
Peter Clark, Isaac Cowhey, Oren Etzioni, Tushar Khot, Ashish Sabharwal, Carissa Schoenick, and Oyvind Tafjord.
\newblock Think you have solved question answering? try arc, the ai2 reasoning challenge.
\newblock \emph{arXiv preprint arXiv:1803.05457}, 2018.

\bibitem[Cobbe et~al.(2021)Cobbe, Kosaraju, Bavarian, Chen, Jun, Kaiser, Plappert, Tworek, Hilton, Nakano, Hesse, and Schulman]{cobbe2021trainingverifierssolvemath}
Karl Cobbe, Vineet Kosaraju, Mohammad Bavarian, Mark Chen, Heewoo Jun, Lukasz Kaiser, Matthias Plappert, Jerry Tworek, Jacob Hilton, Reiichiro Nakano, Christopher Hesse, and John Schulman.
\newblock Training verifiers to solve math word problems, 2021.
\newblock URL \url{https://arxiv.org/abs/2110.14168}.

\bibitem[Dai et~al.(2024)Dai, Deng, Zhao, Xu, Gao, Chen, Li, Zeng, Yu, Wu, et~al.]{dai2024deepseekmoe}
Damai Dai, Chengqi Deng, Chenggang Zhao, RX~Xu, Huazuo Gao, Deli Chen, Jiashi Li, Wangding Zeng, Xingkai Yu, Yu~Wu, et~al.
\newblock Deepseekmoe: Towards ultimate expert specialization in mixture-of-experts language models.
\newblock \emph{arXiv preprint arXiv:2401.06066}, 2024.

\bibitem[Dou et~al.(2024)Dou, Zhou, Liu, Gao, Shen, Xiong, Zhou, Wang, Xi, Fan, Pu, Zhu, Zheng, Gui, Zhang, and Huang]{dou-etal-2024-loramoe}
Shihan Dou, Enyu Zhou, Yan Liu, Songyang Gao, Wei Shen, Limao Xiong, Yuhao Zhou, Xiao Wang, Zhiheng Xi, Xiaoran Fan, Shiliang Pu, Jiang Zhu, Rui Zheng, Tao Gui, Qi~Zhang, and Xuanjing Huang.
\newblock {L}o{RAM}o{E}: Alleviating world knowledge forgetting in large language models via {M}o{E}-style plugin.
\newblock In Lun-Wei Ku, Andre Martins, and Vivek Srikumar (eds.), \emph{Proceedings of the 62nd Annual Meeting of the Association for Computational Linguistics (Volume 1: Long Papers)}, pp.\  1932--1945, Bangkok, Thailand, August 2024. Association for Computational Linguistics.
\newblock \doi{10.18653/v1/2024.acl-long.106}.
\newblock URL \url{https://aclanthology.org/2024.acl-long.106}.

\bibitem[Gao et~al.(2025)Gao, Chen, Rao, Liu, Sun, Zhang, Peng, Guo, and Subrahmanian]{gao-etal-2025-mola}
Chongyang Gao, Kezhen Chen, Jinmeng Rao, Ruibo Liu, Baochen Sun, Yawen Zhang, Daiyi Peng, Xiaoyuan Guo, and Vs~Subrahmanian.
\newblock {M}o{LA}: {M}o{E} {L}o{RA} with layer-wise expert allocation.
\newblock In Luis Chiruzzo, Alan Ritter, and Lu~Wang (eds.), \emph{Findings of the Association for Computational Linguistics: NAACL 2025}, pp.\  5097--5112, Albuquerque, New Mexico, April 2025. Association for Computational Linguistics.
\newblock \doi{10.18653/v1/2025.findings-naacl.284}.
\newblock URL \url{https://aclanthology.org/2025.findings-naacl.284/}.

\bibitem[Han et~al.(2025)Han, Du, Du, Zhou, Wu, Zhang, Zheng, and Han]{han-etal-2025-slim}
Jiayi Han, Liang Du, Hongwei Du, Xiangguo Zhou, Yiwen Wu, Yuanfang Zhang, Weibo Zheng, and Donghong Han.
\newblock {SLIM}: Let {LLM} learn more and forget less with soft {L}o{RA} and identity mixture.
\newblock In Luis Chiruzzo, Alan Ritter, and Lu~Wang (eds.), \emph{Proceedings of the 2025 Conference of the Nations of the Americas Chapter of the Association for Computational Linguistics: Human Language Technologies (Volume 1: Long Papers)}, pp.\  4792--4804, Albuquerque, New Mexico, April 2025. Association for Computational Linguistics.
\newblock \doi{10.18653/v1/2025.naacl-long.246}.
\newblock URL \url{https://aclanthology.org/2025.naacl-long.246/}.

\bibitem[Hayou et~al.(2024)Hayou, Ghosh, and Yu]{hayou2024lora+}
Soufiane Hayou, Nikhil Ghosh, and Bin Yu.
\newblock Lora+ efficient low rank adaptation of large models.
\newblock In \emph{Proceedings of the 41st International Conference on Machine Learning}, pp.\  17783--17806, 2024.

\bibitem[Hu et~al.(2022)Hu, Wallis, Allen-Zhu, Li, Wang, Wang, Chen, et~al.]{hulora}
Edward~J Hu, Phillip Wallis, Zeyuan Allen-Zhu, Yuanzhi Li, Shean Wang, Lu~Wang, Weizhu Chen, et~al.
\newblock Lora: Low-rank adaptation of large language models.
\newblock In \emph{International Conference on Learning Representations}, 2022.

\bibitem[Jiang et~al.(2024)Jiang, Sablayrolles, Roux, Mensch, Savary, Bamford, Chaplot, Casas, Hanna, Bressand, et~al.]{jiang2024mixtral}
Albert~Q Jiang, Alexandre Sablayrolles, Antoine Roux, Arthur Mensch, Blanche Savary, Chris Bamford, Devendra~Singh Chaplot, Diego de~las Casas, Emma~Bou Hanna, Florian Bressand, et~al.
\newblock Mixtral of experts.
\newblock \emph{arXiv preprint arXiv:2401.04088}, 2024.

\bibitem[Jin et~al.(2025)Jin, Zhu, Yuan, and YAN]{jin2025moe}
Peng Jin, Bo~Zhu, Li~Yuan, and Shuicheng YAN.
\newblock Moe++: Accelerating mixture-of-experts methods with zero-computation experts.
\newblock In \emph{The Thirteenth International Conference on Learning Representations}, 2025.
\newblock URL \url{https://openreview.net/forum?id=t7P5BUKcYv}.

\bibitem[Kalajdzievski(2023)]{kalajdzievski2023rank}
Damjan Kalajdzievski.
\newblock A rank stabilization scaling factor for fine-tuning with lora.
\newblock \emph{arXiv preprint arXiv:2312.03732}, 2023.

\bibitem[Li et~al.(2024)Li, Ma, Wang, Cheng, Duan, Zuo, Yang, and Tang]{li2024mixlora}
Dengchun Li, Yingzi Ma, Naizheng Wang, Zhiyuan Cheng, Lei Duan, Jie Zuo, Cal Yang, and Mingjie Tang.
\newblock Mixlora: Enhancing large language models fine-tuning with lora based mixture of experts.
\newblock \emph{arXiv preprint arXiv:2404.15159}, 2024.

\bibitem[Liu et~al.(2024{\natexlab{a}})Liu, Feng, Xue, Wang, Wu, Lu, Zhao, Deng, Zhang, Ruan, et~al.]{liu2024deepseek}
Aixin Liu, Bei Feng, Bing Xue, Bingxuan Wang, Bochao Wu, Chengda Lu, Chenggang Zhao, Chengqi Deng, Chenyu Zhang, Chong Ruan, et~al.
\newblock Deepseek-v3 technical report.
\newblock \emph{arXiv preprint arXiv:2412.19437}, 2024{\natexlab{a}}.

\bibitem[Liu et~al.(2024{\natexlab{b}})Liu, Wang, Yin, Molchanov, Wang, Cheng, and Chen]{liudora}
Shih-yang Liu, Chien-Yi Wang, Hongxu Yin, Pavlo Molchanov, Yu-Chiang~Frank Wang, Kwang-Ting Cheng, and Min-Hung Chen.
\newblock Dora: Weight-decomposed low-rank adaptation.
\newblock In \emph{Forty-first International Conference on Machine Learning}, 2024{\natexlab{b}}.

\bibitem[Mihaylov et~al.(2018)Mihaylov, Clark, Khot, and Sabharwal]{mihaylov2018can}
Todor Mihaylov, Peter Clark, Tushar Khot, and Ashish Sabharwal.
\newblock Can a suit of armor conduct electricity? a new dataset for open book question answering.
\newblock In \emph{Proceedings of the 2018 Conference on Empirical Methods in Natural Language Processing}, pp.\  2381--2391, 2018.

\bibitem[Sakaguchi et~al.(2021)Sakaguchi, Bras, Bhagavatula, and Choi]{sakaguchi2021winogrande}
Keisuke Sakaguchi, Ronan~Le Bras, Chandra Bhagavatula, and Yejin Choi.
\newblock Winogrande: An adversarial winograd schema challenge at scale.
\newblock \emph{Communications of the ACM}, 64\penalty0 (9):\penalty0 99--106, 2021.

\bibitem[Talmor et~al.(2019)Talmor, Herzig, Lourie, and Berant]{talmor2019commonsenseqa}
Alon Talmor, Jonathan Herzig, Nicholas Lourie, and Jonathan Berant.
\newblock Commonsenseqa: A question answering challenge targeting commonsense knowledge.
\newblock In \emph{Proceedings of the 2019 Conference of the North American Chapter of the Association for Computational Linguistics: Human Language Technologies, Volume 1 (Long and Short Papers)}, pp.\  4149--4158, 2019.

\bibitem[vLLM Team()]{vllm}
vLLM Team.
\newblock vllm.
\newblock \url{https://github.com/vllm-project/vllm}.

\bibitem[Wang et~al.(2024)Wang, Cheng, Zhan, Li, Song, and Liu]{wangopenchat}
Guan Wang, Sijie Cheng, Xianyuan Zhan, Xiangang Li, Sen Song, and Yang Liu.
\newblock Openchat: Advancing open-source language models with mixed-quality data.
\newblock In \emph{The Twelfth International Conference on Learning Representations}, 2024.

\bibitem[Yang et~al.(2024)Yang, Yang, Hui, Zheng, Yu, Zhou, Li, Li, Liu, Huang, Dong, Wei, Lin, Tang, Wang, Yang, Tu, Zhang, Ma, Yang, Xu, Zhou, Bai, He, Lin, Dang, Lu, Chen, Yang, Li, Xue, Ni, Zhang, Wang, Peng, Men, Gao, Lin, Wang, Bai, Tan, Zhu, Li, Liu, Ge, Deng, Zhou, Ren, Zhang, Wei, Ren, Liu, Fan, Yao, Zhang, Wan, Chu, Liu, Cui, Zhang, Guo, and Fan]{yang2024qwen2technicalreport}
An~Yang, Baosong Yang, Binyuan Hui, Bo~Zheng, Bowen Yu, Chang Zhou, Chengpeng Li, Chengyuan Li, Dayiheng Liu, Fei Huang, Guanting Dong, Haoran Wei, Huan Lin, Jialong Tang, Jialin Wang, Jian Yang, Jianhong Tu, Jianwei Zhang, Jianxin Ma, Jianxin Yang, Jin Xu, Jingren Zhou, Jinze Bai, Jinzheng He, Junyang Lin, Kai Dang, Keming Lu, Keqin Chen, Kexin Yang, Mei Li, Mingfeng Xue, Na~Ni, Pei Zhang, Peng Wang, Ru~Peng, Rui Men, Ruize Gao, Runji Lin, Shijie Wang, Shuai Bai, Sinan Tan, Tianhang Zhu, Tianhao Li, Tianyu Liu, Wenbin Ge, Xiaodong Deng, Xiaohuan Zhou, Xingzhang Ren, Xinyu Zhang, Xipin Wei, Xuancheng Ren, Xuejing Liu, Yang Fan, Yang Yao, Yichang Zhang, Yu~Wan, Yunfei Chu, Yuqiong Liu, Zeyu Cui, Zhenru Zhang, Zhifang Guo, and Zhihao Fan.
\newblock Qwen2 technical report, 2024.
\newblock URL \url{https://arxiv.org/abs/2407.10671}.

\bibitem[Zellers et~al.(2019)Zellers, Holtzman, Bisk, Farhadi, and Choi]{zellers2019hellaswag}
Rowan Zellers, Ari Holtzman, Yonatan Bisk, Ali Farhadi, and Yejin Choi.
\newblock Hellaswag: Can a machine really finish your sentence?
\newblock In \emph{Proceedings of the 57th Annual Meeting of the Association for Computational Linguistics}, pp.\  4791--4800, 2019.

\end{thebibliography}
\bibliographystyle{iclr2026_conference}

\appendix
\section{Appendix}
\subsection{Details of the involved datasets}

\begin{table}[h!]
\centering
\caption{Summary of datasets used in the experiments.}
\begin{tabular}{lcccc}
\toprule
\textbf{Dataset} & \textbf{Task Type} & \textbf{Train} & \textbf{Dev} & \textbf{Test} \\
\midrule
OpenBookQA (OBQA) & Multiple-choice QA & 4,957 & 500 & 500 \\
BoolQ & Binary QA & 9,400 & 3,200 & 3,200 \\
HellaSwag & Sentence completion & 40,000 & — & 10,000 \\
WinoGrande (Debiased) & Commonsense reasoning & 9,248 & 1,267 & 1,767 \\
CommonsenseQA (CSQA) & Commonsense QA & 9,798 & 1,224 & 1,225 \\
ARC-c & Commonsense reasoning & 1,418 & — & 1,172 \\
ARC-e & Commonsense reasoning & 2,821 & — & 2,376 \\
GSM8K & Mathematics & 7,200 & — & 1,300 \\
HumanEval & Code generation & — & — & 164 \\
Code Alpaca & Code generation & 20,000 & — & — \\
\bottomrule
\end{tabular}
\end{table}

All the mentioned datasets are open-sourced and allow academic use. We report results for the test set when the ground truth is available. Otherwise, we use the dev set.

\subsection{Comparison on hold out datasets  }
We compare the catastrophic forgetting resulting from LoRA and FURINA. N/A represents the original model. We fine-tune the Llama3.2-3B model with different PEFT approaches on the OBQA dataset, and evaluate on the hold-out datasets, GSM8K and MMLU. The result demonstrates that, compared with LoRA, FURINA could significantly mitigate catastrophic forgetting. Note that the ``Rel. Perf. Drop'' is calculated as follows:
\begin{equation}
    \begin{array}{l}
        \text{Rel. Perf. Drop} = \frac{1}{N}\sum\limits_{i=1}^N\frac{\text{ACC}_i^0 - \text{ACC}_i}{\text{ACC}_i^0},
    \end{array}
\end{equation}
in which $\text{ACC}_i^0$ and $\text{ACC}_i$ represent the performance of the original and the fine-tuned model on task $i$, respectively.
\iffalse
\begin{table}[htbp]
  \centering
  \caption{Comparison on out-of-domain dataset. The model is trained on the OBQA dataset.}
    \begin{tabular}{ccccc}
    \toprule
    Method & GSM8K(\%) & MMLU(\%)  & IQuiz(\%) & \makecell[c]{Rel. Perf. Drop (\%,$\downarrow$)} \\
    \midrule
    N/A   & 74.7  & 55.8  & 38.3  & N/A \\
    \midrule
    LoRA  & 60.4 ({\color{red}14.3$\downarrow$})  & 49.7 ({\color{red}6.1$\downarrow$})  & 37.5 ({\color{red}0.8$\downarrow$})  & 10.8 \\
    {\bf FURINA} & 74.5 ({\color{red}0.2$\downarrow$})  & 52.6 ({\color{red}3.1$\downarrow$})  & 36.7 ({\color{red}1.6$\downarrow$})  & 3.4 \\
    {\bf FURINA} w/ rsLoRA & 73.5 ({\color{red} 1.2$\downarrow$}) & 50.6 ({\color{red} 5.2$\downarrow$})  & 40.0 (\textcolor[rgb]{ .306,  .643,  .918}{1.7$\uparrow$})  & 2.2 \\
    {\bf FURINA} w/ LoRA+ & 73.5 ({\color{red}1.2$\downarrow$})  & 51.5 ({\color{red}4.3$\downarrow$}) & 37.5 ({\color{red}0.8$\downarrow$}) & 3.8 \\
    \bottomrule
    \end{tabular}%
  \label{tab:addlabel}%
\end{table}%
\fi
\begin{table}[htbp]
  \centering
  \caption{Comparison on out-of-domain dataset. The model is trained on the OBQA dataset.}
    \begin{tabular}{cccc}
    \toprule
    Method & GSM8K(\%) & MMLU(\%)  & \makecell[c]{Rel. Perf. \\Drop (\%,$\downarrow$)} \\
    \midrule
    N/A   & 74.7  & 55.8  & N/A \\
    \midrule
    LoRA  & 60.4 ({\color{red}14.3$\downarrow$})  & 49.7 ({\color{red}6.1$\downarrow$})  & 10.1 \\
    {\bf FURINA} & 74.5 ({\color{red}0.2$\downarrow$})  & 52.6 ({\color{red}3.1$\downarrow$})  & \textcolor[rgb]{ .306,  .643,  .918}{\bf 2.0} \\
    {\bf FURINA} w/ rsLoRA & 73.5 ({\color{red} 1.2$\downarrow$}) & 50.6 ({\color{red} 5.2$\downarrow$})  & 3.7 \\
    {\bf FURINA} w/ LoRA+ & 73.5 ({\color{red}1.2$\downarrow$})  & 51.5 ({\color{red}4.3$\downarrow$}) & 3.1 \\
    \bottomrule
    \end{tabular}%
  \label{tab:addlabel}%
\end{table}%

\subsection{Inference time of PEFT approaches}
\label{appendix:time}
To evaluate the inference latency of different PEFT methods, we conduct a controlled comparison using a fixed sequence length of 100 input and 100 output tokens. Time to First Token (TTFT) measures the latency until the first output token is generated, while total latency refers to the time cost for generating the entire output sequence. The evaluation is performed across two distinct frameworks to assess both unmerged and merged deployment scenarios:
\begin{enumerate}
\item \textbf{MoE-PEFT Framework}: This framework is used to evaluate standard MoE-LoRA approaches directly. For a fair comparison, single-adapter LoRA methods are also evaluated in this framework without merging the adapters into the backbone model. Our proposed FURINA method is merged into the standard LoRA architecture under this framework.
\item \textbf{vLLM Framework}: This setup represents a production-ready deployment environment. In this scenario, adapters are merged into the backbone model prior to inference, which is only feasible for mergeable PEFT approaches, such as vanilla LoRA and our FURINA. The router-based MoE-LoRA baselines cannot be evaluated in this setting as they are fundamentally unmergeable due to their routing mechanisms.
\end{enumerate}
All prompts and hyperparameters are aligned across the compared approaches. To account for potential variance in time measurements, each test is repeated 5 times per model, with reported results representing the averaged values.
Our results in Tab.~\ref{tab:time} demonstrate that FURINA achieves comparable inference latency to standard LoRA while significantly outperforming conventional MoE-LoRA methods. This confirms that FURINA maintains the efficiency benefits of mergeable PEFT methods while delivering the performance advantages of MoE approaches.
\begin{table}[htbp]
  \centering
  \caption{Inference time of different PEFT approaches on different implementation frameworks. $^{\dag}$ represents FURINA without shared experts.}
    \begin{tabular}{lccccr}
\cmidrule{1-5}    \multicolumn{1}{c}{\multirow{2}[4]{*}{Method}} & \multicolumn{2}{c}{MoE-PEFT} & \multicolumn{2}{c}{vLLM} &  \\
\cmidrule{2-5}          & TTFT(ms)  & Latency(ms) & TTFT(ms)  & Latency(ms) &  \\
\cmidrule{1-5}    \multicolumn{5}{c}{Single Adapter LoRA} &  \\
\cmidrule{1-5}    LoRA  & $\approx$350  & $\approx$3500  & $\approx$10  & $\approx$800  &  \\
    DoRA  & $\approx$400  & $\approx$9000  & $\approx$10  & $\approx$800   &  \\
\cmidrule{1-5}    \multicolumn{5}{c}{Router-based MoE of LoRA} &  \\
\cmidrule{1-5}    MixLoRA & $\approx$550  & $\approx$9000  & N/A   & N/A   &  \\
    LoRAMoE & $\approx$450 & $\approx$6000  & N/A   & N/A   &  \\
    MoLA  & $\approx$700  & $\approx$14500 & N/A   & N/A   &  \\
    SLIM  & $\approx$650  & $\approx$17000 & N/A   & N/A   &  \\
\cmidrule{1-5}    \multicolumn{5}{c}{Router-free MoE of LoRA} &  \\
\cmidrule{1-5}    {\bf FURINA$^{\dag}$ (Ours)} & $\approx$350 & $\approx$3500  & $\approx$10  & $\approx$800   &  \\
    {\bf FURINA (Ours)} & $\approx$350  & $\approx$3500  & $\approx$10  & $\approx$800  &  \\
\cmidrule{1-5}    \end{tabular}%
  \label{tab:time}%
\end{table}%
\vspace{-3mm}
\subsection{Comparison with directly eliminating the routers of standard MoE--LoRA}
To evaluate the effectiveness of the proposed FURINA, we also compare FURINA with directly eliminating the routers during inference. We utilize SLIM as the baseline and merge the adapters as:
\begin{equation}
    \begin{array}{l}
        \textbf{B}=\frac{1}{N}(B_1, B_2, \ldots, B_N), \textbf{A} = (A_1^T, A_2^T, \ldots, A_N^T)^T.
    \end{array}
\end{equation}
We utilize $\frac{1}{N}$ to synthesize the re-weighting operation of the routers. The experiment is conducted on Llama3.2-3B. The results are demonstrated in Tab.~\ref{tab:SLIM-merge}. Directly eliminating the routers from the MoE--LoRA could significantly decrease its performance, even worse than the original model.

\begin{table}[htbp]
    \centering
    \caption{Effect of directly eliminating routers from MoE--LoRA approaches during inference}
    \begin{tabular}{cccccc}
    \toprule
        Method     &   Original Model   &   SLIM   &   SLIM w/ merge   &   {\bf FURINA} \\
    \midrule
        Perf. (\%) &   67.5             &   79.2   &    37.4 ({\color{red}41.8$\downarrow$})           &   {\bf 79.7}     \\
    \bottomrule
    \end{tabular}
    \label{tab:SLIM-merge}
\end{table}
\vspace{-3mm}

\subsection{Activation pattern of LoRA and FURINA}
\begin{figure}[htbp]
    \centering
    \includegraphics[width=1.\linewidth]{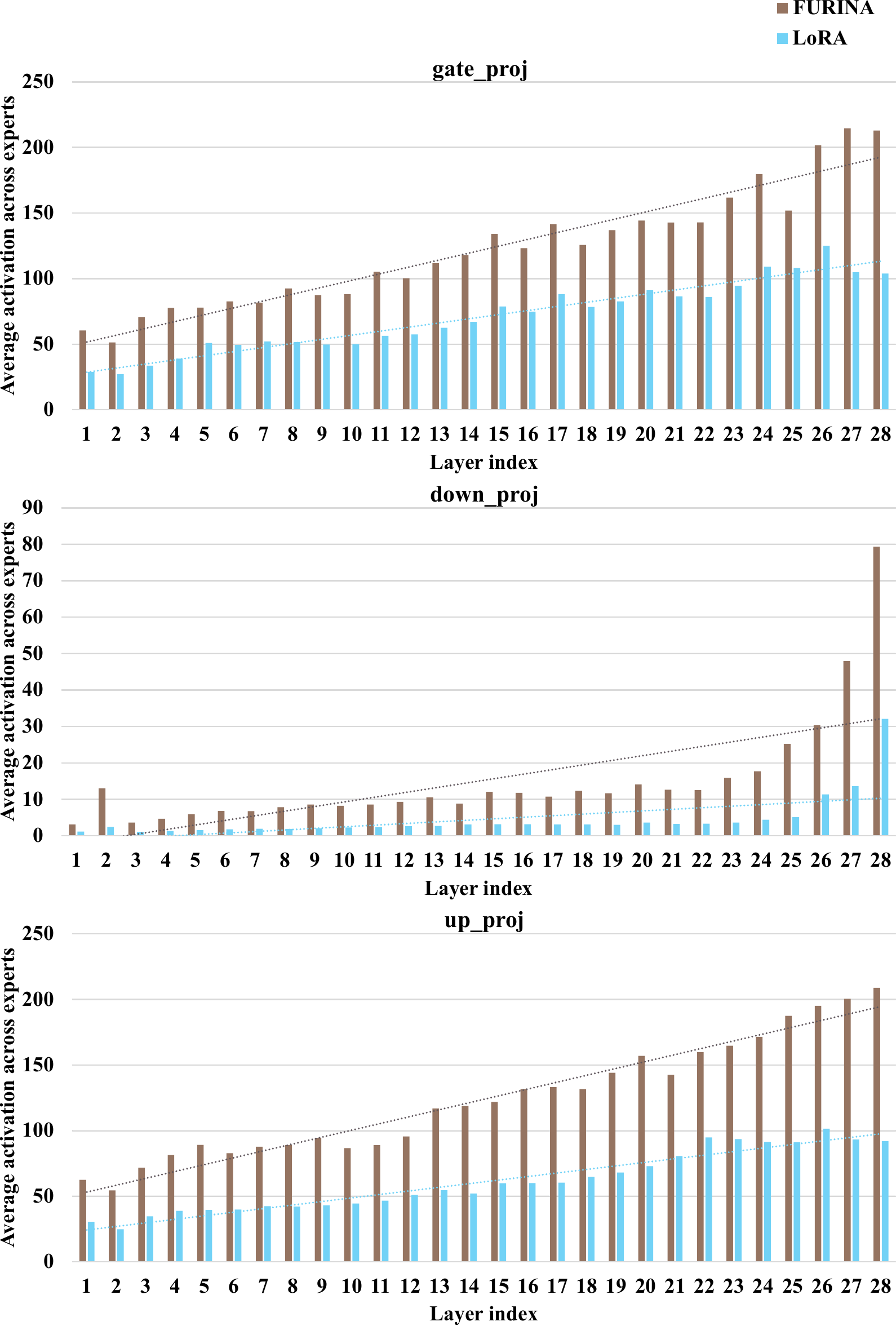}
    \caption{Comparison of activation pattern of FURINA and LoRA across layers}
    \label{fig:vs-lora}
\end{figure}

In Fig.~\ref{fig:vs-lora}, we also compare the activation pattern of FURINA and LoRA. Note that we compare the normalized activation of both approaches on OBQA dataset, with the Llama-3.2-3B model. The result demonstrates that, compared with LoRA, FURINA achieves significantly larger normalized activations, indicating that the proposed approach could effectively increase the angular similarity of input and LoRA weights.

\subsection{Hyper-Parameters of evaluation on EvalScope.}
To validate the fine-tuned models on GSM8K and HumanEval datasets with the EvalScope framework in the main paper, the hyperparameters are set as in Tab.~\ref{tab:hyper}.
\begin{table}[htbp]
\caption{Hyper-parameters for evaluating GSM8K and HumanEval}
    \centering
    \begin{tabular}{ccccc}
    \toprule
        \#max tokens   &   temperature  & \# shots \\
    \midrule
        2048           &   0.0          & 0\\
    \bottomrule
    \end{tabular}
    \label{tab:hyper}
\end{table}

\end{document}